\documentclass[conference]{IEEEtran}
\IEEEoverridecommandlockouts
\usepackage{cite}
\usepackage{amsmath,amssymb,amsfonts}
\usepackage[english]{babel}
\usepackage{graphicx}
\usepackage{textcomp}
\usepackage{xcolor}

\usepackage[utf8]{inputenc}
\usepackage[T1]{fontenc}
\usepackage{float}
\usepackage{multirow}
\usepackage{stfloats}
\usepackage{amsmath, amssymb}
\usepackage{xcolor, colortbl}
\usepackage{tabularray}
\usepackage{subcaption}
\usepackage{newfloat}
\usepackage{tikz} 
\usepackage{url}
\usepackage{algorithm}
\usepackage{algpseudocode}

\usepackage{url}

\usepackage{draftwatermark}
\SetWatermarkText{ACCEPTED \\MANUSCRIPT}
\SetWatermarkScale{.6}
\SetWatermarkColor[gray]{0.9}

\begin{document}

\title{Decoding Market Emotion from Blockchain Activity: A Data-Driven Sentiment Classifier\thanks{This manuscript has been accepted for presentation at the IEEE International Symposium on Computers and Communications (ISCC 2026).}
}



\author{\IEEEauthorblockN{
Arthur G. Bubolz\IEEEauthorrefmark{1}, 
Abreu Quevedo\IEEEauthorrefmark{1}, 
Giancarlo Lucca\IEEEauthorrefmark{1},
Rafael A. Berri\IEEEauthorrefmark{1},
Eduardo Borges\IEEEauthorrefmark{1},
Bruno L. Dalmazo\IEEEauthorrefmark{1}
} 
\IEEEauthorblockA{\IEEEauthorrefmark{1}Federal University of Rio Grande - FURG, Brazil}%
Computing Sciences Center - C3 \\
E-mail: \{arthurgomesbubolz, abreu\_rg, giancarlo.lucca, rafaelberri, eduardoborges, dalmazo\}@furg.br
}

\maketitle

\begin{abstract}

The growing use of Bitcoin as a decentralized digital asset and investment tool has sparked strong interest in understanding its market behavior. This study presents a new approach to analyze Bitcoin market sentiment by combining on-chain and financial data with social media posts. Unlike models that aim to predict prices, this work focuses on explaining market sentiment using blockchain transactions, historical price data of Bitcoin and daily Twitter sentiment classifications. The method merges sentiment trends with on-chain and financial metrics, normalized into a dataset for detailed market analysis. Multiple machine learning models were tested using cross-validation, with Gradient Boosting (XGBoost) emerging as the most reliable for classifying sentiment, achieving an average F1-score of about 0.84. SHAP (SHapley Additive exPlanations), a game theory–based method for model interpretability, was used to quantify the contribution of on-chain features to the model’s predictions, improving transparency. The results indicate that this data combination yields meaningful predictive signals and insights, supporting data-driven cryptocurrency analysis and future improvements with deep learning.

\end{abstract}


\section{Introduction}

Over the last two decades, the growing adoption of cryptographic assets, particularly Bitcoin, has captured the attention of investors and researchers, consolidating itself as a significant phenomenon in global financial markets. The evolution of this market has generated challenges and opportunities in the analysis of these assets, which have become emerging issues as the cryptocurrency market has advanced.

Among the various approaches used for these objectives, the analysis of market sentiments from social network information, as discussed in \cite{11326523}, stands out, a technique that often presents interesting results dealing with the influence of different social networks in relation to market behaviour. At the same time, a relevant aspect is the analysis of on-chain data, which uses the information available on the Bitcoin network to describe market dynamics, as illustrated in the study by \cite{jagannath2021chain}. However, it is observed that, for the most part, the analyzes focus on one approach or another, always with a forecasting perspective, whether trend \cite{10.1007/978-3-031-96997-3_3} or market quotation \cite{mcnally2018predicting, test-exsy}.

This study aims to integrate two distinct approaches that, despite employing different methodologies, share the same goal. However, unlike predictive studies, such as those discussed by \cite{chevallier2021possible}, which address the difficulties in predicting dynamic prices, the focus of this study is explanatory, seeking to understand market sentiment, analyzing tweet data and their respective sentimental classifications based on a set of retrieved data. From this analysis, we seek to summarize daily sentiment based on tweets, while training a model using on-chain and financial data to classify market sentiment.

This process aims to investigate whether information from the Bitcoin network has a direct correlation with what is discussed on social network posts, providing a more integrated view of market sentiment. The idea is to provide greater security to investors, creating an alternative indicator that reflects the general mood of the market, offering relevant insights for analysts and investors. For the evaluation, the cross-validation technique is applied to machine learning algorithms to analyze on-chain data to extract a clear perception of the prevailing sentiment in the market.

Before presenting the technical details, the key contributions of this study are summarized as follows:
(i) integrating blockchain data with sentiment information from social network posts to create a unified dataset that helps explain market behavior;
(ii) providing a straightforward method to understand market sentiment using Bitcoin network data without relying on complex external sources;
(iii) investigating how blockchain activity may reflect or influence online discussions, highlighting potential connections between network behavior and public opinion;
(iv) proposing a simpler approach to track market sentiment without accessing social network APIs or processing large volumes of text, while offering model explainability through SHAP values to enhance the interpretability of results.

This article is organized as follows: Section~\ref{sec:relatedWork} presents the related work that supports this research, highlighting their approaches and contributions. In Section~\ref{sec:methodology}, the methodology adopted, the development environment, the data used, the pre-processing strategies and the evaluation metrics are detailed. Section~\ref{sec:results} discusses the results obtained from the experiments carried out. Finally, Section~\ref{sec:conclusions} presents the conclusions of this study, its limitations and suggestions for future work.

\section{Related work}
\label{sec:relatedWork}

This section aims to present an overview of the main existing approaches that use on-chain variables to classify Bitcoin price behavior. Work that explores on-chain analysis within the context of machine learning will be discussed, with an emphasis on the respective predictive techniques applied. The intention is to identify the most relevant methods used to classify market behavior, which had the on-chain variables most considered in these studies.

In the context of predicting Bitcoin volatility through the analysis of on-chain data combined with Tweet publications, especially from the @whale\_alert twitter account \cite{azamjon2023forecasting}, discusses the use of reinforcement learning, more specifically Q-learning, to build a high-precision predictive model to classify variations in the price of Bitcoin.

\begin{table*}[hb]
\centering
\caption{Summary of the main characteristics of the related work}
\begin{tabular}{p{5cm}cccccccc}
\hline \hline
\textbf{Main features} & {\cite{ali2023ensemble}} & {\cite{casella2023predicting}} & {\cite{kim2022deep}} & {\cite{jagannath2021self}} & {\cite{brinckman2019techniques}} & {\cite{bello2023lld}} & {\cite{badruddoja2022making}} & {\cite{azamjon2023forecasting}} \\ \hline \hline
Uses technical data & \checkmark & \checkmark & \checkmark & \checkmark & \checkmark & \checkmark & \checkmark & \checkmark  \\ 
Uses textual data & & & & & & & & \checkmark \\ 
Regressive Approach & & & & \checkmark & \checkmark & \checkmark & & \checkmark \\ 
Classification Approach & \checkmark & \checkmark & \checkmark & & \checkmark & & \checkmark & \checkmark \\ 
Short Forecast & \checkmark & \checkmark & \checkmark & \checkmark & \checkmark & \checkmark & \checkmark & \checkmark \\ 
Long Forecast & & & & & & & & \checkmark \\ 
Traditional ML Algorithms & & & & & \checkmark & \checkmark & \checkmark & \\ 
Deep Learning & \checkmark & \checkmark & \checkmark & \checkmark & \checkmark & \checkmark & & \\ 
Reinforcement Learning & & & & & & & & \checkmark \\ \hline \hline
\end{tabular}
\end{table*}

Regarding security, \cite{badruddoja2022making} presents innovative approaches to analyzing smart contracts on Ethereum, applying machine learning algorithms and utilizing unique features such as using integers instead of floating point numbers to calculate probabilities. Although this methodology has proven to be effective and scalable, its practical applicability still depends on greater integration with the final objective of the research.

Still focusing on security, specifically anomaly detection using on-chain data, \cite{bello2023lld} proposed a model to detect low-latency pump-and-dump schemes by analyzing 1-minute charts and using an LSTM autoencoder to predict cryptocurrency prices and spot anomalies in real time. While effective, the limited sample size may reduce its broader applicability.

Recent advances in blockchain data analysis and price prediction have been explored in studies like \cite{brinckman2019techniques}, which highlight challenges and propose solutions. A key area involves extracting and analyzing Ethereum blockchain data, facing difficulties in identifying outliers and suspicious smart contracts. Extracting transaction data and mapping contract interfaces is complex due to the diversity and detail of blockchain information. Additionally, scalability, data management, and smart contract complexity still hinder effective representation of relevant information.

Self-adaptive deep learning models such as jSO-LSTM have been explored in \cite{jagannath2021self} to predict Bitcoin prices, taking into account multiple on-chain metrics. Data normalization and the selection of highly correlated variables were crucial to the model's performance, which outperformed conventional methods. Furthermore, frameworks such as SAM-LSTM, explored by \cite{kim2022deep}, use attention mechanisms combined with LSTM to predict cryptocurrency prices, emphasizing the importance of detecting change points to adapt the model to previously unseen variations.

Predicting the phases of the cryptocurrency market, especially to guide investment strategies, has also been addressed in studies such as \cite{casella2023predicting}, which use on-chain data. Predictive models such as SARIMA, LSTM and CNN were compared, showing the advantage of deep learning methods over traditional approaches, although incorporating predictions of actual market movement remains an area for exploration.

Finally, cryptocurrency price prediction using deep learning ensemble models, which combine CNNs and RNNs, was highlighted in \cite{ali2023ensemble} as a promising approach. These models can handle the volatility and complex dependencies of blockchain data, surpassing existing methodologies and paving the way for applications in modeling the price of cryptographic assets.

The mentioned studies address the use of on-chain data for various purposes, such as cryptocurrency price forecasting, anomaly detection, and market manipulation scheme identification. While presenting innovative and technically sophisticated approaches, they often focus on direct price predictions or isolated historical trend analysis. Even when normalizing data, they treat context uniformly and disaggregated, limiting the ability to provide a more strategic and contextual view of market movements.

This study addresses gaps in previous research by proposing a new indicator to measure market sentiment. The indicator uses social network data verified with on-chain information, removing the need to gather data from multiple sources. Instead of predicting future cryptocurrency prices, the model determines whether the market is optimistic or pessimistic. Using on-chain data as a decision-support tool, combined with advanced machine learning techniques, we aim to identify whether the market is in a bullish or bearish phase, thereby assisting investment decision-making. The main characteristics of the related works are summarized in Table 1.

\section{Proposal and Adopted Methodology} \label{sec:methodology}

The present study proposes the development of a predictive model that integrates on-chain data with sentiment analysis from social network posts, with the aim of exploring the relationship between behavior recorded on the blockchain and opinion trends expressed in tweets. The main idea is to build an alternative indicator that does not depend on multiple sources with data of varying nature to identify market sentiment. With just some specific data and in a more simplified way, this indicator aims to describe the general sentiment of the market, offering greater security to the investor and more autonomy, allowing them to calculate the metrics in a simple way and make more robust decisions.

On-chain data is used as an input variable (X), representing information such as transactions and other metrics extracted from the blockchain. On the other hand, the output variable (Y) is composed of the summary of the feelings of the tweets extracted from Kaggle, where each tweet is classified according to its emotional charge (positive, negative or neutral). To ensure that both datasets are compatible and optimized for modeling, both are normalized to fit the requirements of the predictive model using appropriate scaling techniques.

\begin{figure}[H]
\centering
\begin{center}
\includegraphics[width=\linewidth]{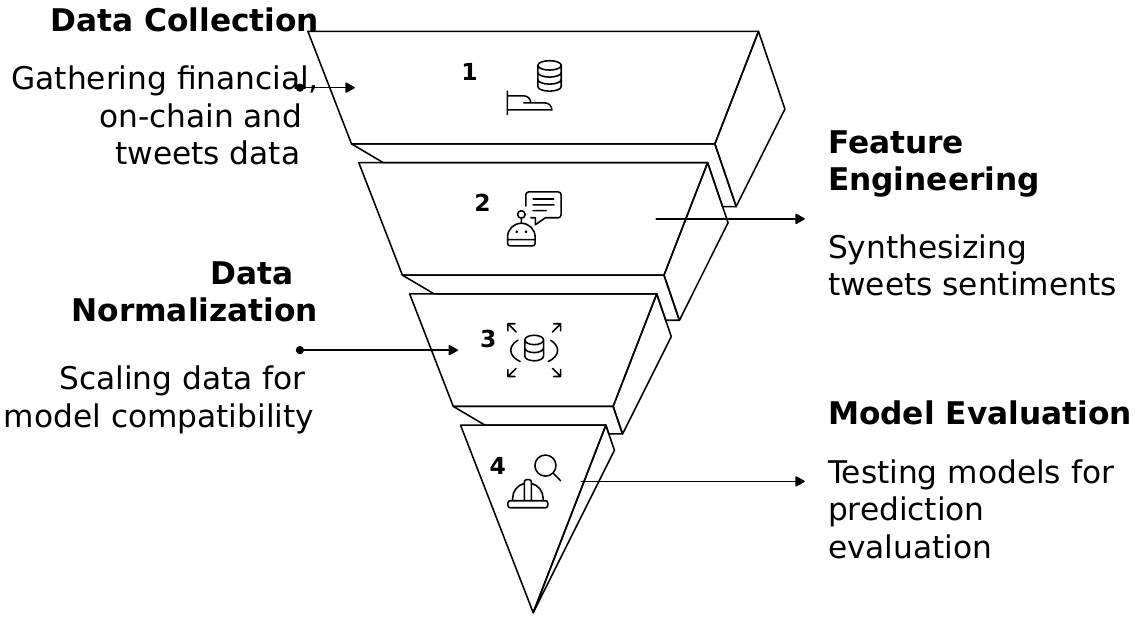}
\caption{\label{meth_diagram}Proposed methodology diagram}
\end{center}
\end{figure}

The initial step (1 in Figure~\ref{meth_diagram}) consists of collecting financial data from Bitcoin price information on a daily timeframe, also on-chain data, which includes detailed information about the blockchain, such as transactions and metrics associated with a specific period. Simultaneously, tweet data is extracted, containing associated sentiments, which reflect market opinions and perceptions. This step aims to integrate blockchain information with users' sentimental trends on social network posts, specifically in the context of the cryptocurrency market.

In the second stage (2 in Figure~\ref{meth_diagram}), a feature engineering process is carried out, where each day's tweets are synthesized to identify the predominant sentiment. By analyzing the most recurrent class of sentiment among tweets on a given day, market sentiment is classified as positive or negative. These values are subsequently aggregated and joined to the blockchain data through an inner join, creating a single set of data that unites the two sources of information based on their registration date.

In the third stage (3 in Figure~\ref{meth_diagram}), the unified data undergoes normalization using the \textit{MinMax Scaler} technique, applied at intervals of 30 records. This ensures the data is on a uniform scale, enabling the model to identify patterns more efficiently and accurately, facilitating the analysis of interactions between on-chain data and market sentiment.

In the fourth stage (4 in Figure~\ref{meth_diagram}), several machine learning models are evaluated using the cross-validation technique, with the aim of identifying the model that can best classify market sentiment fluctuations based on technical variables extracted from the blockchain. In summary, this work investigates whether combining Bitcoin network data with social-network features can accurately summarize the market behavior, as collective sentiment may indicate crypto-market volatility.

Based on the studies discussed in the Related Work section, it is evident that on-chain data has been extensively explored for tasks such as price forecasting, volatility prediction, anomaly detection, and market phase identification, often through deep learning models or reinforcement learning techniques \cite{bello2023lld}. While these approaches demonstrate strong predictive capabilities, they predominantly treat on-chain variables as inputs to forecast future price movements or detect abnormal behavior, with limited emphasis on explaining market sentiment itself. In contrast, this study proposes a different methodological perspective by repositioning on-chain data as the primary explanatory source for market sentiment, rather than as an auxiliary signal for price prediction.

Unlike approaches that directly combine on-chain metrics with textual features from social networks \cite{azamjon2023forecasting} or rely on complex deep learning architectures for long- or short-term forecasting, the proposed methodology uses sentiment extracted from social network posts solely as a reference label to represent the daily market mood. The classification model is trained on on-chain and market-derived technical features, allowing the analysis to focus on whether blockchain activity alone reflects collective sentiment trends, integrating data aggregation, temporal normalization, and cross-validated machine learning classification, this approach provides a simplified yet robust framework that emphasizes interpretability and analytical transparency offering an explanatory alternative that aligns on-chain behavior with market sentiment, rather than future price outcomes.

\subsection{Data Collection}

As illustrated in Algorithm~\ref{alg:data_pipeline_highlevel_en}, the data collection implements a comprehensive pipeline for collecting, transforming, and normalizing Bitcoin network data, emphasizing the consolidation of daily metrics to enable consistent trend analysis.

The Blockchair platform\footnote{$https://gz.blockchair.com/bitcoin/blocks/$} is used to access Bitcoin network block records. Through a GET request, the code retrieves the HTML content of the records page and parses it to extract links to daily block files from 2013 to 2026. The function performing this process is defined in line 1 of Algorithm~\ref{alg:data_pipeline_highlevel_en}. After downloading, records are processed to calculate metrics such as average block weight, difficulty, reward in USD, transaction count, inputs, outputs, witnesses, and total fees.

The normalization of metrics is carried out carefully to reflect the characteristics of each type of data correctly. Values such as weight, difficulty and reward are treated as averages, representing a centralized measure of each day's data, ideal for understanding the average behavior of each metric over time. Values such as transactions, entries, exits, witnesses and fees are added together, as each transaction, entry, exit and witness represents a unit of activity, and the sum of these variables throughout the day offers a more accurate view of the total amount of activities and costs associated with the network.

\begin{algorithm}[htbp]
\caption{Data Collection and Aggregation Pipeline}
\label{alg:data_pipeline_highlevel_en}
\begin{algorithmic}[1]
\Require List of Data File URLs (Links)
\Ensure Consolidated DataFrame ($DF\_Result$) with Aggregated Metrics

\Statex \textbf{Function} $\mathrm{Filter\_and\_Download}(\mathrm{Links})$
\State Filter Links excluding initial data years
\State Create or verify the \texttt{downloads} directory
\ForAll{link in Filtered\_Links}
    \State Download link to directory \texttt{downloads/}
\EndFor
\State \Return List\_of\_Downloaded\_Files

\Statex

\Statex \textbf{Function} $\mathrm{Process\_File}(\mathrm{Raw\_DF})$
\State Calculate period-based aggregated metrics from Raw\_DF
\State \Return DF\_Row

\Statex

\Statex \textbf{Main Process}

\State $DF\_Result \gets$ Empty DataFrame()
\State $Files \gets \mathrm{Filter\_and\_Download}(\mathrm{Links})$

\ForAll{file in Files}

    \State $Raw\_DF \gets \mathrm{LoadCSV}(file)$

    \If{$Raw\_DF$ is not empty}
        \State $DF\_Row \gets \mathrm{Process\_File}(Raw\_DF)$
        \State $DF\_Result \gets DF\_Result \cup DF\_Row$
        \Comment{Append the aggregated row}
    \EndIf

\EndFor

\State \Return $DF\_Result$

\end{algorithmic}
\end{algorithm}

The \textit{df\_to\_row} function, defined in line~7 of Algorithm~\ref{alg:data_pipeline_highlevel_en}, transforms the data into a format suitable for time series analysis by converting each block DataFrame into a single row containing the normalized statistics for that day. The code then aggregates these rows into a final DataFrame called \textit{df\_result}, which includes exception handling for empty files or read errors. This process is defined from line~9 to the end of Algorithm~\ref{alg:data_pipeline_highlevel_en}.
Finally, the records are organized in chronological order to maintain the temporal coherence of the data, ensuring that the sequence of events is preserved for subsequent analyses.

\subsection{Feature Engineering}

Initially, tweet data for sentiment analysis was extracted from the Kaggle dataset\footnote{$https://www.kaggle.com/datasets/gauravduttakiit/bitcoin-tweets-16m-tweets-with-sentiment-tagged$}. The goal was to determine daily market sentiment by aggregating individual tweet-level classifications, assigning each day the majority class of positive and negative tweets.

Although the dataset includes a neutral class, exploratory analysis showed that neutral tweets rarely dominated daily counts. In practice, they were the majority on very few days, creating statistical instability and limiting the reliability of a three-class approach.

From an interpretative view, the neutral class does not provide a clear signal of market optimism or pessimism, which is the focus of this study. Including it would increase noise and reduce the model’s ability to capture meaningful sentiment dynamics. Therefore, the analysis used a binary sentiment formulation (positive vs. negative), ensuring a more stable learning process, clearer interpretability, and alignment with the study’s explanatory goals.
After labeling daily sentiment, the data were joined with on-chain information by date, integrating blockchain metrics with market sentiment classifications to create a unified dataset for further analysis. 
The dataset resulting from these treatments is publicly available in the research file 
repository\footnote{$https://github.com/arthur-bubolz/Sentiment-Index-2026$}, 
ensuring full transparency and allowing other researchers to reproduce the results.

\subsection{Data Normalization}

The data preparation followed a structured approach to make it suitable for predictive analysis. On-chain data and daily sentiment, classified by timestamp, were joined and reorganized to create a more intuitive temporal structure. Instead of the traditional past-to-present order, data were arranged from the most recent to the earliest day, reflecting the practical workflow where analysis often starts with recent information.

Next, the data were normalized to ensure comparability across variables. A block-based normalization was applied, dividing the dataset into fixed-size segments and scaling each block independently, preventing variables with larger ranges from dominating the results, which is critical for machine learning models sensitive to scale differences.

Each block was transformed using \textit{MinMaxScaler} to the range 0–1, preserving the relative distribution within blocks. Blocks were processed sequentially and then concatenated into a complete dataset ready for predictive modeling.

Finally, the dependent variable, representing daily sentiment classification (positive or negative), was aligned with the input data. This ensured integrated predictions without temporal mismatches, maintaining data quality and providing a solid basis for models exploring relationships between on-chain data and sentiments from social media posts.

\subsection{Model Evaluation}

The model was evaluated using cross-validation with various classification algorithms, adopting the F1-score as the main performance metric.

Although deep learning has been widely used for cryptocurrency tasks such as price forecasting and volatility prediction, its use here is intentionally limited. The goal is not to maximize predictive accuracy with complex models, but to check if classical and tree-based algorithms can capture meaningful patterns from on-chain data that reflect market sentiment.

The F1-score was chosen to ensure the model can properly discriminate between the two classes. As the harmonic mean of precision and recall, it balances the ability to correctly identify positive instances and to capture all relevant positives, accounting for both false positives and false negatives—essential for this study.
Finally, the best decision-tree model from cross-validation is selected and trained separately to evaluate its performance more accurately. This model is also preferred for later SHAP analysis, as its segmented, interpretable structure facilitates variable explainability.

\section{Evaluation and Results} \label{sec:results}

To evaluate the models, a 5-fold cross-validation procedure was adopted, using the F1-score as the primary metric. This choice ensures a balanced assessment of precision and recall, which is essential given the importance of both classes (0 and 1). The evaluation results are presented in the following image:

\begin{figure}[H]
\centering
\begin{center}
\includegraphics[width=\linewidth]{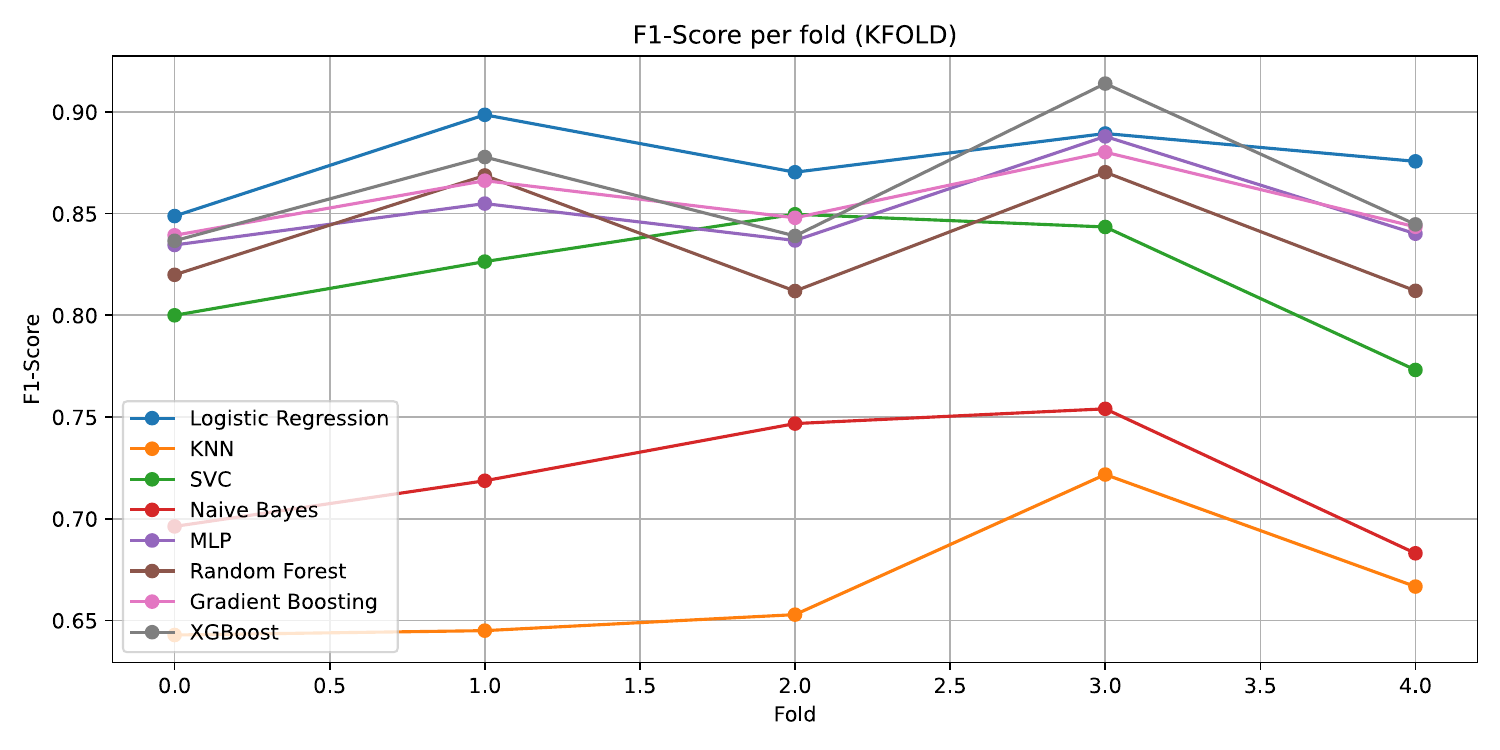}
\caption{\label{cross_results}F1-Score results in Cross-validation per Fold}
\end{center}
\end{figure}

The diagram illustrates the F1-score values per fold of the cross-validation, providing a detailed view of the model's performance in each of the data subsets used in the process. From the analysis of the results, we observed that the XGBoost model was the most stable tree-based algorithm, although it did not stand out in all performance metrics in each fold. XGBoost showed consistent behavior and was the model that obtained the best results in terms of stability, which is a crucial aspect in practical applications. 

\begin{figure}[hb]
\vspace{-2mm}
\centering
\begin{center}
\includegraphics[width=\linewidth]{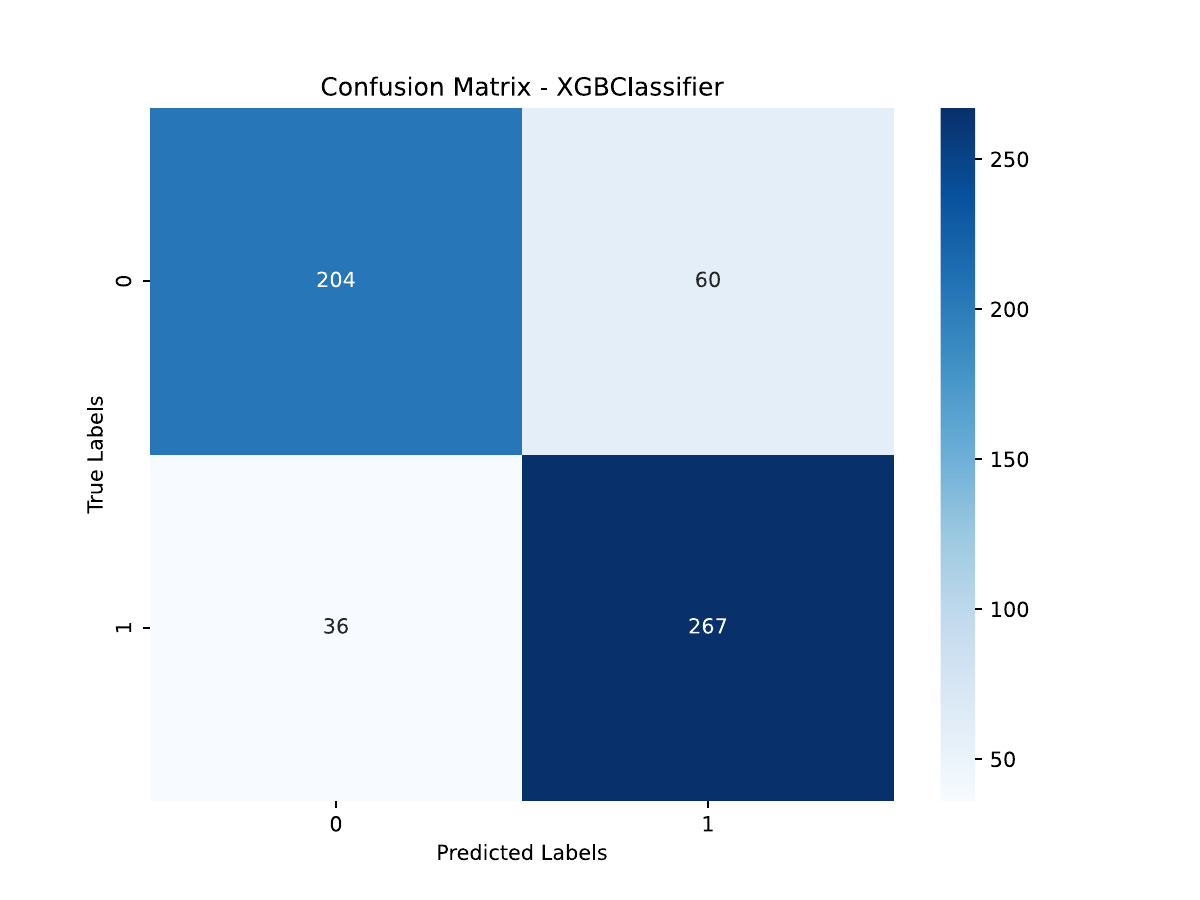}
\caption{\label{matrix_conf}Confusion Matrix of XGBoost}
\end{center}
\end{figure}

To verify the robustness of the cross-validation results, additional training was performed using the XGBoost model, since it was the best-performing tree-based algorithm and is well suited for analyzing feature importance with SHAP\footnote{$https://shap.readthedocs.io/en/latest/$}
. The data was split into 70\% for training and 30\% for testing, allowing a more rigorous evaluation of model performance. The confusion matrix from this split is shown in Figure~\ref{matrix_conf}.

Furthermore, the model's key performance metrics are presented in the table below, offering a detailed view into the effectiveness of the XGBoost model in terms of accuracy, precision, recall and F1-score demonstrated in Table ~\ref{tab:metricas}.

\begin{table}[ht]
\vspace{2mm}
    \centering
    \caption{XGBoost Model Performance Metrics}
    \label{tab:metricas}
    \renewcommand{\arraystretch}{1.3}
    \begin{tabular}{p{2cm}|p{2cm}<{\raggedleft\arraybackslash}}
        \hline \hline
        \textbf{Metric} & \textbf{Score} \\ \hline
        Accuracy & 0.8307 \\ 
        Precision & 0.8165 \\ 
        Recall & 0.8812 \\ 
        F1-Score & 0.8476 \\ 
        \hline \hline
    \end{tabular}
\end{table}

The results obtained, reflected in the performance metrics, indicate that the model achieved robust and effective performance, with good accuracy in class classification. The high F1-Score, in particular, suggests that the model is balanced, managing to classify both the positive and negative classes consistently. This positive performance is highly beneficial as it lays a solid foundation for the development of predictive tools that can be applied in real-world scenarios, such as analyzing market sentiments and forecasting economic trends. 

To complete the analysis, the Permutation Importance technique was used to evaluate the relevance of the features to the model. The metric chosen to measure the model's performance was the F1-score, as it offers a good balance between precision and recall and has already been used previously to evaluate models in general.
The XGBoost model was used for this analysis, maintaining the previous hyperparameters of the previously trained model: \textit{use\_label\_encoder=False, eval\_metric='mlogloss', random\_state=42}. 

The result of this process provided us with a detailed view of the relative importance of the variables, allowing us to identify which features have the greatest impact on the model's prediction, and are therefore most relevant to its performance, guiding possible adjustments to the model, such as feature selection or re-evaluation of the pre-processing strategy, in addition to providing valuable insights into the model's behavior in relation to the data.

\begin{figure}[ht]
\centering
\begin{center}
\includegraphics[width=\linewidth]{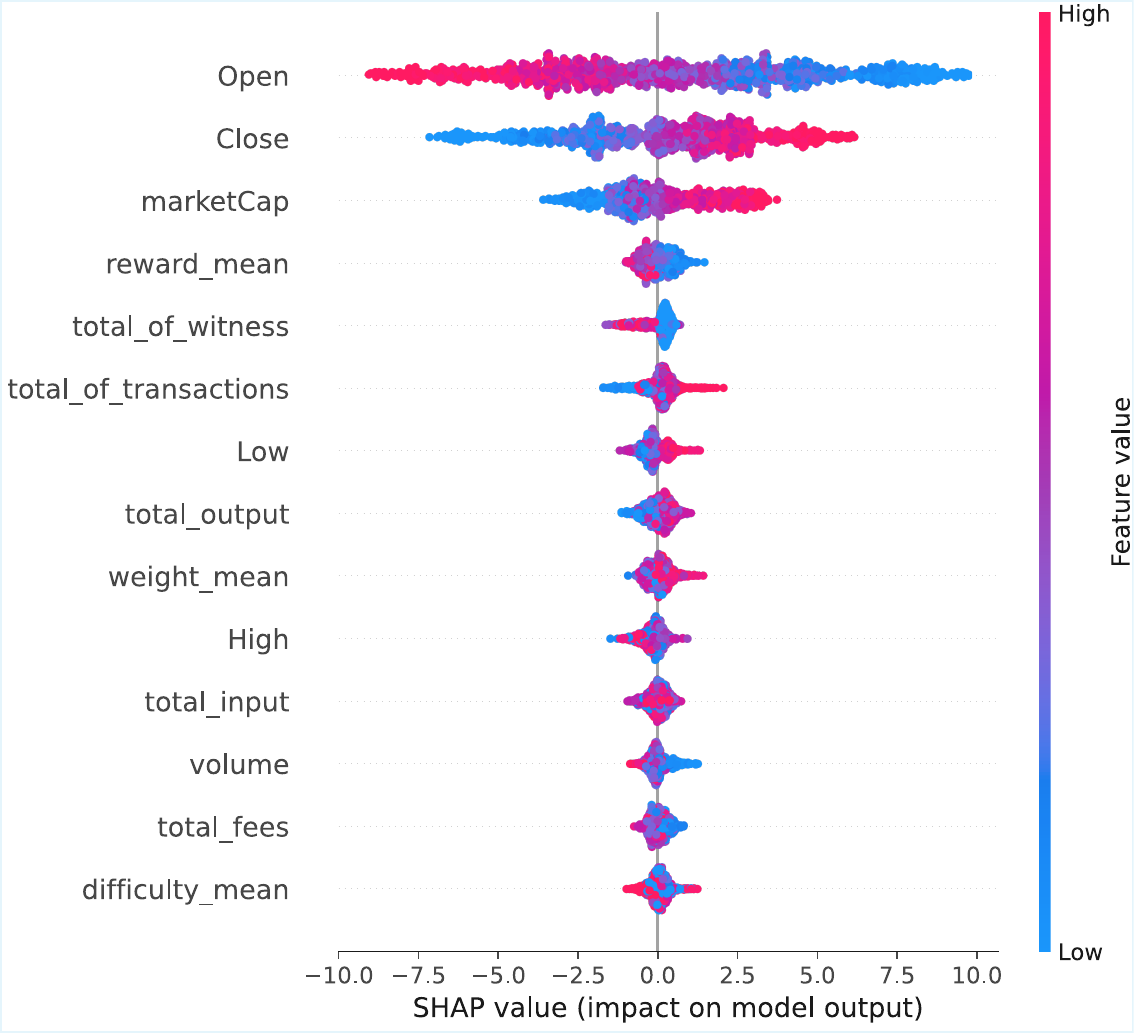}
\caption{\label{shap_values}SHAP values}
\end{center}
\end{figure}

Based on the results presented on Figure~\ref{shap_values}, we can observe that the \textit{Open} variable presented the highest importance score, which makes it the most relevant variable in the model. Indicating that the opening price is directly related to daily market conditions and can reflect financial behavior at key moments, such as at the beginning of a new trading period. Furthermore, other variables with a strong financial relationship, such as \textit{Close} and \textit{marketCap}, also presented significant scores, reinforcing the idea that financial variables have a great influence on the model's prediction.

Alternatively, variables related to on-chain behavior, such as \textit{reward\_mean}, also have an important influence. This indicates that factors directly associated with price movements are more strongly correlated with the predictions.

The results suggest that, for the model, the relevance of financial variables is significant, due to the reflection of market behavior and financial conditions in a clear and objective way. At the same time, on-chain variables also have an important impact, especially those that are directly related to transactional activity on the blockchain. In summary, variables connected to actual market transactions show higher relevance in the model.

\section{Conclusions} \label{sec:conclusions}

This article presented a study on Bitcoin market sentiment, integrating on-chain data and social network information. After the calibration and adjustment process, the results indicated that the model satisfactorily associated on-chain variables with daily sentiment, achieving significant performance in the proposed task. The XGBoost algorithm stood out as the best-performing tree-based model, with an average F1-score of approximately 0.84, demonstrating the effectiveness of the approach. This promising performance suggests that it is possible to develop a robust indicator based on on-chain data and social network information, without requiring real-time access to social network APIs or continuous text processing pipelines, to provide a reliable perspective on market sentiment through the analysis of historical and sentiment data.

In terms of future perspectives, a possible line of investigation would be the use of deep learning techniques in conjunction with XGBoost, through ensemble methods or other approaches, in order to improve the already successful model. This combination could explore the complexities of the data in greater depth, enhancing the accuracy of predictions. Furthermore, the XGBoost model, with its ability to describe community sentiment well, could be integrated with other machine learning models to help as an auxiliary signal for downstream predictive tasks, creating an even more powerful tool for analysis and decision-making in the financial context.

\bibliographystyle{IEEEtran}
\bibliography{sbc-template.bib}

\end{document}